\documentclass[11pt,a4paper]{article}
\usepackage[hyperref]{eacl2021}
\usepackage{times}
\usepackage{latexsym}

\usepackage{microtype}

\usepackage{graphicx}
\usepackage{amsmath}
\usepackage{booktabs}

\aclfinalcopy 

\title{Syntax-BERT: Improving Pre-trained Transformers with Syntax Trees}
  
\author{\textbf{Jiangang Bai\textsuperscript{\rm 1,\thanks{~~The work was done when the author visited Microsoft Research Asia.}}, Yujing Wang\textsuperscript{\rm 1,2,\thanks{~~Corresponding Author}}, Yiren Chen\textsuperscript{\rm 1}, Yaming Yang\textsuperscript{\rm 2}}\\
\textbf{Jing Bai\textsuperscript{\rm 2}, Jing Yu\textsuperscript{\rm 3}, Yunhai Tong\textsuperscript{\rm 1}}\\
\textsuperscript{\rm 1}Peking University, Beijing, China\\
\textsuperscript{\rm 2}Microsoft Research Asia, Beijing, China \\
\textsuperscript{\rm 3}Institute of Information Engineering, Chinese Academy of Sciences, Beijing, China\\
\texttt{\small \{pku\_bjg,yujwang,yrchen92,yhtong\}@pku.edu.cn}\\
\texttt{\small \{yujwang,yayaming,jbai\}@microsoft.com}\\
\texttt{\small yujing02@iie.ac.cn}
}

\date{}

\begin{document}
\maketitle
\begin{abstract}
Pre-trained language models like BERT achieve superior performances in various NLP tasks without explicit consideration of syntactic information. Meanwhile, syntactic information has been proved to be crucial for the success of NLP applications. However, how to incorporate the syntax trees effectively and efficiently into pre-trained Transformers is still unsettled. In this paper, we address this problem by proposing a novel framework named Syntax-BERT. 
This framework works in a plug-and-play mode and is applicable to an arbitrary pre-trained checkpoint based on Transformer architecture. Experiments on various datasets of natural language understanding verify the effectiveness of syntax trees and achieve consistent improvement over multiple pre-trained models, including BERT, RoBERTa, and T5. At the same time, we also made our experiment code public\footnote{https://github.com/nkh2235/SyntaxBERT.git}.
\end{abstract}

\section{Introduction}
Pre-trained language models like BERT~\cite{devlin2018bert}, RoBERTa~\cite{liu2020roberta} and T5~\cite{raffel2019exploring} become popular in recent years and achieve outstanding performances in various NLP benchmarks. These models often choose a Transformer architecture largely owing to its attractive scalability. Studies~\cite{hewitt2019structural, jawahar-etal-2019-bert} have shown that a pre-trained transformer is able to capture certain syntactic information implicitly by learning from sufficient examples. However, there is still a big gap between the syntactic structures implicitly learned and the golden syntax trees created by human experts.

On the other hand, syntax tree is a useful prior for NLP-oriented neural networks~\cite{kiperwasser2018scheduled}.
For example, Tree-LSTM~\cite{tai2015improved} extends the sequential architecture of LSTM to a tree-structured network. Linguistically-informed self-attention (LISA)~\cite{strubell2018linguistically} proposes a multi-task learning framework for semantic role labeling, which incorporates syntactic knowledge into Transformer by training one attention head to be attended to its parent in a syntax tree. In addition, \citet{Nguyen2020TreestructuredAW} integrate tree-structured attention in Transformer with hierarchical accumulation guided by the syntax tree. 

Although there are numerous works on syntax-enhanced LSTM and Transformer models, none of the previous works have addressed the usefulness of syntax-trees in the pre-training context. It is straight-forward to ask: \textit{it is still helpful to leverage syntax trees explicitly in the pre-training context?} If the answer is yes, \textit{can we ingest syntax trees into a pre-trained checkpoint efficiently without training from scratch for a specific downstream application?} This is an appealing feature in practice because pre-training from scratch is a huge waste of energy and time. 

In this paper, we propose Syntax-BERT to tackle the raised questions. Unlike a standard BERT, which has a complete self-attention typology, we decompose the self-attention network into multiple sub-networks according to the tree structure. Each sub-network encapsulates one relationship from the syntax trees, including ancestor, offspring, and sibling relationships with different hops. All sub-networks share the same parameters with the pre-trained network, so they can be learned collaboratively and inherited directly from an existing checkpoint. To select the task-oriented relationships automatically, we further adopt a topical attention layer to calculate the relative importance of syntactic representations generated by different sub-networks. Finally, the customized representation is calculated by weighted summation of all sub-networks.

We conduct extensive experiments to verify the effectiveness of Syntax-BERT framework on various NLP tasks, including sentiment classification, natural language inference, and other tasks in the GLUE benchmark. Experimental results show that Syntax-BERT outperforms vanilla BERT models and LISA-enhanced models consistently with multiple model backbones, including BERT, RoBERTa, and T5. Specifically, it boosts the overall score of GLUE benchmark from 86.3 to 86.8 for T5-Large~\cite{raffel2019exploring} checkpoint, which is already trained on a huge amount of data. This improvement is convincing since only a few extra parameters are introduced to the model. 

Our \textbf{major contributions} are as follows: 
\begin{itemize}
    \item To the best of our knowledge, Syntax-BERT is one of the first attempts to demonstrate the usefulness of syntax trees in pre-trained language models. It works efficiently in a plug-and-play fashion for an existing checkpoint without the need for pre-training from scratch.
    \item To integrate syntax trees into pre-trained Transformers, we propose a novel method that decomposes self-attention networks into different aspects and adopts topical attention for customized aggregation. As shown in the ablation study, this design benefits from syntactic structures effectively while retaining pre-trained knowledge to the largest extent.
    \item Syntax-BERT shows consistent improvement over multiple pre-trained backbone models with comparable model capacities. It can be combined with LISA to achieve further enhancement, indicating that these two algorithms are complementary to each other.
\end{itemize}

\begin{figure*}[t]
    \centering
    \includegraphics[width=\linewidth]{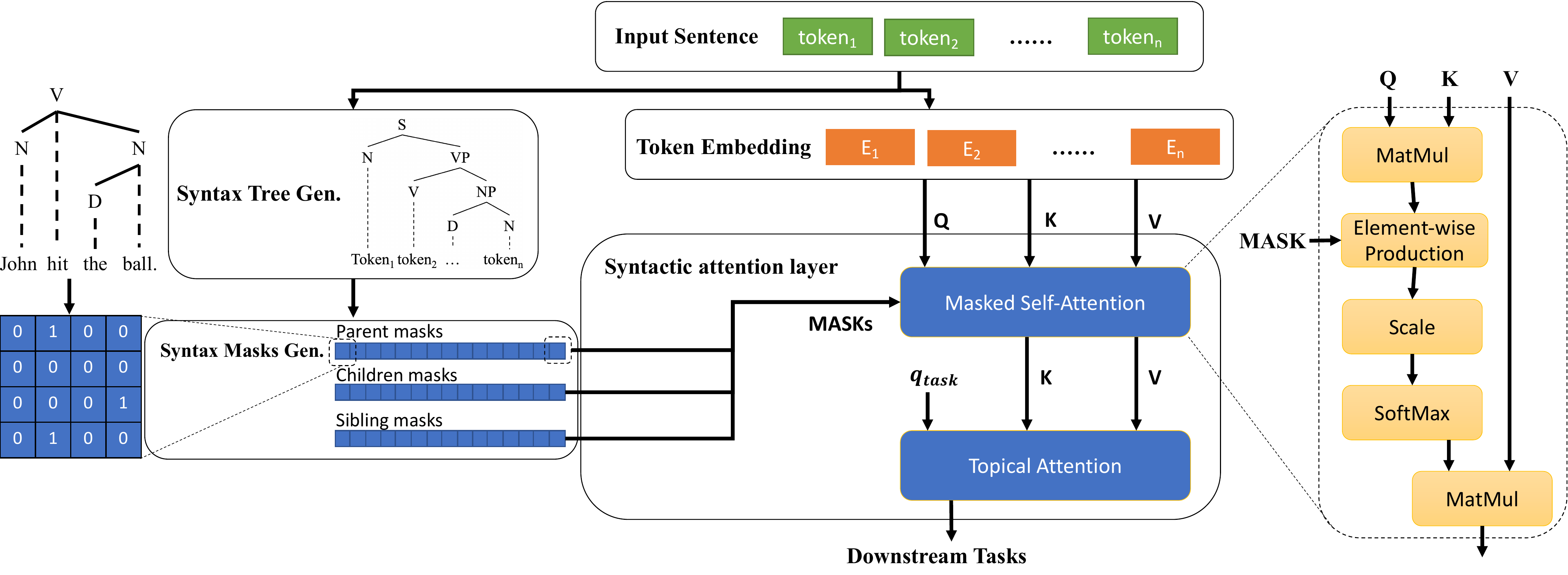}
    \caption{{\small The Overall Architecture of Syntax-BERT. Note that the leftmost part shows an example of syntax tree and its corresponding parent syntax mask ($d = 1$).}}
    \label{fig:architecture}
\end{figure*}

\section{Related Work}
\subsection{Pre-trained language models}
Recently, pre-trained language models have received significant attention from the natural language processing community. Many excellent pre-trained language models are proposed, such as BERT, RoBERTa and T5.
Transformer~\cite{vaswani2017attention} is a typical architecture for pre-training language models, which is based on the self-attention mechanism and is much more efficient than RNNs.
BERT~\cite{devlin2018bert} is a representative work that trains a large language model on the free text and then fine-tunes it on specific downstream tasks separately. 
BERT is pre-trained on two auxiliary pre-training tasks, Masked Language Model (MLM) and Next Sentence Prediction (NSP).
RoBERTa~\cite{liu2020roberta} is an improved variant of BERT which utilizes dynamic masks.
In RoBERTa, the NSP task is cancelled, but the full-sentence mechanism is considered. At the same time, the size of RoBERTa's training data ($\sim$160GB) is ten times the size of BERT's training data.
Moreover, \citet{raffel2019exploring} explore the effectiveness of multiple transfer learning techniques and apply these insights at scale to create a new model T5 (Text to Text Transfer Transformer).
With T5, they reform all NLP tasks into a unified text-to-text format where the input and output are always text strings. This is in contrast to BERT-style models that only output either a class label or an input span. 

\subsection{Syntax-aware models}
Syntax is a crucial prior for NLP-oriented neural network models. Along this direction, a range of interesting approaches have been proposed, like Tree-LSTM~\cite{tai2015improved}, PECNN~\cite{yang2016position}, SDP-LSTM~\cite{xu2015classifying}, Supervised Treebank Conversion~\cite{jiang2018supervised}, PRPN~\cite{shen2018neural}, and ON-LSTM~\cite{shen2018ordered}.

Recent works also investigate syntactic knowledge in the context of Transformer, which are more related to this paper.
For instance, Syntax-Infused Transformer~\cite{sundararaman2019syntax} feeds the extra syntactic features into the Transformer models explicitly, but it only considers simple syntactic features and does not provide a generic solution to incorporate tree-structured knowledge. 
\citet{strubell2018linguistically} present a neural network model named LISA (Linguistically-Informed Self-Attention) that learns multi-head self-attention in a multi-task learning framework consisting of dependency parsing, part-of-speech tagging, predicate detection, and semantic role labeling. They also show that golden syntax trees can dramatically improve the performance of semantic role labeling. Moreover, \citet{Nguyen2020TreestructuredAW} propose a hierarchical accumulation approach to encode parse tree structures into self-attention mechanism. However, these approaches are designed for training a Transformer from scratch without benefiting from pre-trained checkpoints. Instead, our framework works in a plug-and-play mode and retains the pre-trained knowledge as much as possible for downstream applications. Concurrent to our work, \citet{sachan2020syntax} investigate popular strategies for incorporating dependency structures into pre-trained language models, revealing essential design decisions are necessary for strong performances. 
In addition, \citet{hewitt2019structural} design two sets of probes to determine whether the embedded space can be converted into syntactic information space through a linear transformation. It gives the evaluation metrics to examine how much syntactic information is included in a model.

\section{Syntax-BERT}
\label{sec:method}
Syntax-BERT is a variant of pre-trained Transformer models, which changes the flow of information in a standard BERT network via a syntax-aware self-attention mechanism. First, the overall architecture of Syntax-BERT is presented in Section \ref{architecture}. Then, we introduce the construction of syntax trees and corresponding masks in Section \ref{syntax_mask}. The details of syntactic attention layers will be described in Section \ref{attention}. 

\subsection{Architecture}
\label{architecture}

As mentioned earlier, one limitation of vanilla Transformer is that it simply uses a fully-connected topology of tokens in the pre-trained self-attention layer. Although the self-attention mechanism automatically calculates a relevance score for each token pair, it still suffers from optimization and over-fitting problems, especially when the training data is limited. Some previous works have tried to induce syntactic structure explicitly into self-attention. For instance, in Linguistically-Informed Self-Attention (LISA)~\cite{strubell2018linguistically}, syntax tree is incorporated by training one attention head to be attended to the parent of each token. However, other structural features such as siblings and children are discarded in the model. Moreover, it can not distinguish the usefulness of multiple syntactic features while largely retain the knowledge from a pre-trained checkpoint.

Syntax-BERT is designed to incorporate grammatical and syntactic knowledge as prior in the self-attention layer and support fine-grained adaptation for different downstream tasks. Specifically, it generates a bunch of sub-networks based on sparse masks reflecting different relationships and distances of tokens in a syntax tree. Intuitively, the tokens inside a sub-network often semantically related to each other, resulting in a topical representation. Therefore, we can adopt a topical attention layer to aggregate task-oriented representations from different sub-networks.

The overall architecture of Syntax-BERT is illustrated in Figure~\ref{fig:architecture}.
As shown in the left part of this figure, we generate syntax masks for the input sentence in two steps. First, the input sentence is converted into the corresponding tree structure by a syntax parser. Second, we extract a bunch of syntax-related masks according to different features incorporated in the syntax tree. Next, the sentence is embedded similar to a standard BERT (token + positional + field embedding) and served as input to the self-attention layer. Each self-attention layer in the Syntax-BERT is composed of two kinds of attention modules, namely \textit{masked self-attention} and \textit{topical attention}. In a \textit{masked self-attention} module, we apply syntactic masks to the fully-connected topology, generating topical sub-networks that share parameters with each other. Furthermore, the representations from different sub-networks are aggregated through a \textit{topical attention} module so that the task-related knowledge can be distilled to the final representation vector. 

\subsection{Masks induced by syntax tree}
\label{syntax_mask}

Generically, a syntax tree is an ordered, rooted tree that represents the syntactic structure of a sentence according to some context-free grammar. It can be defined abstractly as $T=\{R, \mathcal{N}, \mathcal{E}\}$, where $R$ is the root of syntax tree, $\mathcal{N}$ and $\mathcal{E}$ stands for node set and edge set respectively. The most commonly-used syntax trees are constituency trees~\cite{chen2014fast} and dependency trees~\cite{zhu2013fast}, and we use both of them in our experiments unless notified.

To utilize the knowledge in a syntax tree effectively, we introduce syntax-based sub-network typologies in the self-attention layer to guide the model. Each sub-network shares the same model parameters with the global pre-trained self-attention layer, while each sub-network reflects a specific aspect of the syntax tree. This procedure can be easily implemented by multiple masks applied to the complete graph topology. 

Without loss of generality, we design three categories of masks reflecting different aspects of a tree structure, namely \textit{parent mask}, \textit{child mask}, and \textit{sibling mask}. For a pairwise inference task that contains a pair of sentences as input, we also apply another mask, i.e., \textit{pairwise mask}, to capture the inter-sentence attention. Moreover, the distances between nodes (tokens) in a tree incorporate semantic relatedness.
Starting from a node $A$, along the edges of a syntax tree, the minimum number of edges required to reach another node $B$ can be regarded as the distance between $A$ and $B$, written as $dist(A,B)$. We create fine-grained masks according to the distance between two nodes to enable customized aggregation of task-oriented knowledge. 

Mathematically, a mask can be denoted by $M \in \{0, 1\}^{n \times n}$, where $M_{i,j} \in \{0, 1\}$ denotes if there is a connection from token $i$ to token $j$, and $n$ is the number of tokens in the current sentence.

In the \textbf{parent mask} with certain distance $d$, we have $M^{p}_{i,j,d}=1$ if and only if the node $i$ is the parent or ancestor of node $j$, at the same time $dist(i, j) = d$. Otherwise, the value will be set as zero. 

In the \textbf{child mask} with certain distance $d$, we have $M^{c}_{i,j,d}=1$ if and only if the node $i$ is the child or offspring of node $j$, at the same time $dist(i, j) = d$. In other words, node $j$ is the parent or ancestor of node $i$.

In the \textbf{sibling mask} with certain distance $d$, we have $M^{s}_{i,j,d}=1$ if and only if we can find their lowest common ancestor and $dist(i, j) = d$. Note that if two nodes are in the same sentence, we can always find the lowest common ancestor, but the value should be zero if the corresponding nodes come from different sentences (in pairwise inference tasks). 

The \textbf{pairwise mask} captures the interaction of multiple sentences in a pairwise inference task. We have $M^{pair}_{i,j}=1$ if and only if both node $i$ and $j$ are from different sentences. we do not consider the distances in-between as the nodes are from different trees.

\subsection{Syntactic attention layers}
\label{attention}
A block of Syntax-BERT contains two kinds of attention modules: \textit{masked self-attention} and \textit{topical attention}. The operations in a \textit{masked self-attention} are similar to a standard self-attention except that we have sparse network connections as defined in the masks. The \textit{masked self-attention} can be formulated as an element-wise multiplication of dot-product attention and its corresponding mask: 
\begin{equation}
\begin{aligned}
& MaskAtt(Q, K, V, M) = \sigma(\frac{QK^\top\odot M}{\sqrt{d}})V \\
& A_{i,j} = MaskAtt(HW_i^Q,HW_i^K,HW_i^V, M_j) \\
& H_j = (A_{1,j} \oplus A_{2,j} \oplus ... \oplus A_{k,j})W^O, j \in 1,...,m
\end{aligned}
\end{equation}

where $Q$, $K$, $V$ represent for the matrix of query, key and value respectively, which can be calculated by the input representation $H$. $M$ represents for the matrix of syntax mask and $\odot$ denotes an operator for element-wise production; $\sigma$ stands for softmax operator; $A_{i,j}$ denotes the attention-based representation obtained by the $i^{th}$ head and $j^{th}$ sub-network; $W_i^Q$, $W_i^K$ and $W_i^V$ represent for the parameters for linear projections; $M_j$ denotes the mask for the $j^{th} $sub-network; and $H_j$ denotes the corresponding output representation.

The output representations from different sub-networks embody knowledge from different syntactic and semantic aspects. Therefore, we leverage another attention layer, named \textit{topical attention} to perform a fine-grained aggregation of these representations. The most distinct part of a \textit{topical attention} is that $q_{task}$ is a trainable query vector for task-specific embedding. Thus, the \textit{topical attention} layer is able to emphasize task-oriented knowledge captured by numerous sub-networks.
\begin{equation}
\begin{aligned}
& TopicAtt(q_{task}, K, V) = \sigma(\frac{q_{task}K^\top}{\sqrt{d}})V \\
& H^O =TopicAtt(q_{task}, HW^K, HW^V)
\end{aligned}
\end{equation}
where $d$ denotes the size of hidden dimension, $q_{task} \in R^{1 \times d}$ is a task-related learnable query embedding vector; $\sigma$ stands for the softmax operator; $H=(H_1, H_2, ..., H_m)^\top \in R^{m \times d}$ is the output representation collected by multiple sub-networks; $W^K$ and $W^V$ are parameters in the feed-forward operations; and $H^O$ stands for the final text representation.

\section{Experiments}
\label{sec:length}
\label{experiments}
\begin{table}[t]
    \renewcommand\arraystretch{1.0}
    \centering
    \scalebox{0.85}{
    {\small
    \begin{tabular}{ccccc}
    \toprule
     \textbf{Task} & \textbf{\#Train} & \textbf{\#Dev} & \textbf{\#Test} & \textbf{\#Class} \\ 
     \midrule
     SST-1 & 8,544 & 1,101 & 2,210 & 5    \\
     SST-2 &  6,920 & 873 & 1,822 & 2    \\
     SNLI & 549,367 & 9,842 & 9,824 & 3 \\
     MNLI & 392,703 & 9,816/9,833 & 9,797/9,848 & 3  \\
     CoLA & 8,551 & 1,042 & 1,064 & 2  \\
     MRPC & 3,669 & 409 & 1,726 & 2  \\
     STS-B & 5,750 & 1,501 & 1,380 & * \\
     QQP & 363,871 & 40,432 & 390,965 & 2  \\
     QNLI & 104,744 & 5,464 & 5,464 & 2  \\
     RTE & 2,491 & 278 & 3,001 & 2 \\
     WNLI & 636 & 72 & 147 & 2 \\
     \bottomrule
    \end{tabular}
    }
    }
    \caption{\small{Dataset Statistics: the character `/' seperate MNLI-m and MNLI-mm, `*' represents for the regression task.}}
    \label{tab:Statistics}
\end{table}

First, we run experiments on the Stanford Sentiment Treebank (SST) dataset~\cite{socher2013recursive} in Section \ref{sec:SST}, which is designed to study the syntactic and semantic compositionality of sentiment classification. Second, in Section \ref{sec:NLI}, we evaluate the performance of Syntax-BERT on two natural language inference datasets: SNLI and MNLI. Then, more empirical results on the GLUE benchmark and a comprehensive ablation study will be presented in Section \ref{sec:GLUE} and \ref{sec:Ablation} respectively. At last, we present the analysis of the structural probes in Section \ref{sec:prob}. 

The statistics of all datasets adopted in this paper are summarized in Table \ref{tab:Statistics}. For each dataset, we optimize the hyper-parameters of Syntax-BERT through grid search on the validation data. Detailed settings can be found in the appendix. In our experiments, we set the maximum value of $dist(A, B)$ in a syntax tree as 15 and use both dependency and constituency trees unless specified. Thus, we have totally 90 ($15 \times 3 \times 2$) sub-networks for single-sentence tasks and 92 ($(15 \times 3 + 1) \times 2$) sub-networks for pairwise inference tasks. 
We adopt Transformer~\cite{vaswani2017attention}, BERT-Base, BERT-Large~\cite{devlin2018bert}, RoBERTa-Base, RoBERTa-Large~\cite{liu2020roberta} and T5-Large~\cite{raffel2019exploring} as backbone models and perform syntax-aware fine-tuning on them. We also compare with LISA (Linguistically-Informed Self-Attention)~\cite{strubell2018linguistically}, a state-of-the-art method that incorporates linguistic knowledge into self-attention operations. Specifically, LISA~\cite{strubell2018linguistically} adopt an additional attention head to learn the syntactic dependency in the tree structure, and the parameters of this additional head are initialized randomly.

\subsection{Stanford Sentiment Treebank}
\label{sec:SST}
The SST dataset contains more than 10,000 sentences collected from movie reviews from the \textit{rottentomatoes.com} website.
The corresponding constituency trees for review sentences are contained in the dataset, where each intermediate node in a tree represents a phrase. All phrases are labeled to one of five fine-grained categories of sentiment polarity. SST-2 is a binary classification task.
We follow a common setting that utilizes all phrases with lengths larger than 3 as training samples, and only full sentences will be used in the validation and testing phase. The hyper parameters for each model are selected by grid search and listed in the appendix. 
We compare Syntax-BERT with vanilla baselines and LISA-enhanced models. The results are listed in Table \ref{tab:SST}.
As shown in the table, our model achieves 4.8 and 4.9 absolute points improvements respectively against the vanilla Transformer with comparable parameter size. By combining our framework with LISA, the results can be further boosted obviously. This indicates that our mechanism is somewhat complementary to LISA. LISA captures the syntactic information through an additional attention head, whereas our framework incorporates syntactic dependencies into original pre-trained attention heads and increases the sparsity of the network. We can see that \textit{Syntax-Transformer + LISA} performs the best among all settings, and similar trends are demonstrated on the BERT-Base and BERT-Large checkpoints.

\begin{table}[t]
 \centering
 \scalebox{0.9}{
 {\small
  \begin{tabular}{l|c|c}
  \toprule
    \textbf{Model} & \textbf{SST-1} & \textbf{SST-2}  \\ 
    \midrule
    Transformer & 48.4 & 86.2 \\
    LISA-Transformer & 52.2 & 89.1 \\
    \textbf{Syntax-Transformer (Ours)} & 52.7 & 90.1 \\
    \textbf{Syntax-Transformer + LISA (Ours)} & \textbf{53.2} & \textbf{91.1}\\
    \midrule
    BERT-Base & 53.7 & 93.5 \\
    LISA-BERT-Base & 54.2 & 93.7 \\
    \textbf{Syntax-BERT-Base (Ours)} & 54.4 & 94.0 \\
    \textbf{Syntax-BERT-Base + LISA (Ours)} & \textbf{54.5} & \textbf{94.4} \\
    \midrule
    BERT-Large & 54.8 & 94.9 \\
    LISA-BERT-Large & 55.0 & 95.9 \\
    \textbf{Syntax-BERT-Large (Ours)} & 55.3 & 96.1 \\
    \textbf{Syntax-BERT-Large + LISA (Ours)} & \textbf{55.5} & \textbf{96.4} \\
    \bottomrule
  \end{tabular}
 }
}
\caption{\small{Comparison with SOTA models on SST dataset.} 
}
\label{tab:SST}
\end{table}

\subsection{Natural Language Inference}
\label{sec:NLI}

\begin{table}[t]
    \begin{center}
    \scalebox{0.9}{
    {\small
    \begin{tabular}{l|c|c}
    \toprule
    \textbf{Model} & \textbf{SNLI} & \textbf{MNLI} \\
    \midrule
    Transformer & 84.9 & 71.4 \\
    LISA-Transformer & 86.1 & 73.7 \\
    \textbf{Syntax-Transformer (Ours)} & 86.8 & 74.1 \\
    \textbf{Syntax-Transformer + LISA (Ours)} & \textbf{87.0} & \textbf{74.5}\\
    \midrule
    BERT-Base & 87.0 & 84.3 \\
    LISA-BERT-base & 87.4 & 84.7 \\
    \textbf{Syntax-BERT-Base (Ours)} & 87.7 & \textbf{84.9} \\
    \textbf{Syntax-BERT-Base + LISA (Ours)} & \textbf{87.8} & \textbf{84.9}\\
    \midrule
    BERT-Large & 88.4 & 86.8\\
    LISA-BERT-Large & 88.8 & 86.8 \\
    \textbf{Syntax-BERT-Large (Ours)} & 88.9 & 86.7 \\
    \textbf{Syntax-BERT-Large + LISA (Ours)} & \textbf{89.0} & \textbf{87.0}\\
    
    \bottomrule
    \end{tabular}
    }
    }
    \end{center}
    \caption{\small{Comparison with SOTA models on NLI datasets.}}
\label{tab:NLI}
\end{table}

{\small
\begin{table*}[t]
    \scalebox{0.85}{
    \centering
    \small
    \begin{tabular}{lcccccccccc}
    \toprule
     \textbf{Model} & \textbf{Avg} & \textbf{CoLA} & \textbf{SST-2} & \textbf{MRPC} & \textbf{STS-B} & \textbf{QQP} & \textbf{MNLI-m/-mm} & \textbf{QNLI} & \textbf{RTE} & \textbf{WNLI}\\ 
     \midrule
     Transformer & 66.1 & 31.3 & 83.9 & 81.7/68.6 & 73.6/70.2 & 65.6/84.4 & 72.3/71.4 & 80.3 & 58.0 & 65.1 \\
     \textbf{Syntax-Transformer (Ours)} & \textbf{68.8} & \textbf{36.6} & \textbf{86.4} & \textbf{81.8/69.0} & \textbf{74.0/72.3} & \textbf{65.5/84.9} & \textbf{72.5/71.2} & \textbf{81.0} & \textbf{56.7} & 65.1 \\
     \midrule
     BERT-Base & 77.4 & 51.7 & 93.5 & 87.2/82.1 & 86.7/85.4 & 71.1/89.0 & 84.3/83.7 & 90.4 & 67.2 & 65.1\\
     \textbf{Syntax-BERT-Base (Ours)} & \textbf{78.5} & \textbf{54.1} & \textbf{94.0} & \textbf{89.2/86.0} & \textbf{88.1/86.7} & \textbf{72.0/89.6} & \textbf{84.9/84.6} & \textbf{91.1} & \textbf{68.9} & 65.1\\
     \midrule
     BERT-Large & 80.5 & 60.5 & 94.9 & 89.3/85.4 & 87.6/86.5 & 72.1/89.3 & \textbf{86.8/85.9} & 92.7 & 70.1 & 65.1  \\
     \textbf{Syntax-BERT-Large (Ours)} & \textbf{81.8} & \textbf{61.9} & \textbf{96.1} & \textbf{92.0/88.9} & \textbf{89.6/88.5} & \textbf{72.4/89.5} & 86.7/86.6 & \textbf{92.8} & \textbf{74.7}	& 65.1  \\
     \midrule
     RoBERTa-Base & 80.8 & 57.1 & 95.4 & 90.8/89.3 & 88.0/87.4 & 72.5/89.6 & 86.3/86.2 & 92.2 & 73.8 & 65.1  \\
     \textbf{Syntax-RoBERTa-Base (Ours)} & \textbf{82.1} & \textbf{63.3} & \textbf{96.1} & \textbf{91.4/88.5} & \textbf{89.9/88.3} & \textbf{73.5/88.5} & \textbf{87.8/85.7} & \textbf{94.3} & \textbf{81.2} & 65.1 \\
     \midrule
     RoBERTa-Large & 83.9 & 63.8 & 96.3 & 91.0/89.4 & 72.9/90.2 & 72.7/90.1 & 89.5/89.7 & 94.2 & 84.2 & 65.1  \\
     \textbf{Syntax-RoBERTa-Large (Ours)} & \textbf{84.7} & \textbf{64.3} & \textbf{96.9} & \textbf{92.5/90.1} & \textbf{91.6/91.4} & \textbf{73.1/89.8} & \textbf{90.2/90.0} & \textbf{94.5} & \textbf{85.0} & 65.1 \\
     \midrule
     T5-Large & 86.3 & 61.1 & 96.1 & 92.2/88.7 & 90.0/89.2 & 74.1/89.9 & 89.7/89.6 & 94.8 & 87.0 & 65.1  \\
     \textbf{Syntax-T5-Large (Ours)} & \textbf{86.8} & \textbf{62.9} & \textbf{97.2} & \textbf{92.7/90.6} & \textbf{91.3/90.7} & \textbf{74.3/90.1} & \textbf{91.2/90.5} & \textbf{95.2} & \textbf{89.6} & 65.1 \\
     \bottomrule
    \end{tabular}
}
    \caption{\small{Comparison with state-of-the-art models without pre-training on GLUE benchmark.}}
    \label{tab:glue}
\end{table*}
}

The Natural Language Inference (NLI) task requires a model to identify the semantic relationship (entailment, contradiction, or neutral) between a premise sentence and the corresponding hypothesis sentence. In our experiments, we use two datasets for evaluation, namely SNLI~\cite{bowman2015large} and MNLI~\cite{williams2018broad}. We utilize the Stanford parser~\cite{klein2003accurate} to generate constituency and dependency trees for the input sentences. The MNLI dataset has two separate sets for evaluation (matched set and mismatched set), and we report the average evaluation score of these two sets. 

The test accuracies on SNLI and MNLI datasets are shown in Table \ref{tab:NLI}.
The syntactic prior information helps the Transformer to perform much better on the NLI tasks. The accuracies on the SNLI and MNLI datasets have been improved by 1.9 and 2.7, respectively, by applying our framework to a vanilla Transformer. The LISA-enhanced transformer can also outperform vanilla transformer on NLI tasks, but the accuracy improvement is not as large as Syntax-Transformer.
When the backbone model is BERT-Base or BERT-Large, consistent conclusions can be drawn from the experimental results. It is worth noting that the syntax-enhanced models for BERT-large do not show much gain based on the vanilla counterparts. This may because BERT-Large already captures sufficient knowledge for NLI tasks in the pre-training phase. 

\subsection{GLUE Benchmark}
\label{sec:GLUE}

The GLUE benchmark~\cite{wang2018glue} offers a collection of tools for evaluating the performance of models.
It contains single-sentence classification tasks (CoLA and SST-2), similarity and paraphrase tasks (MRPC, QQP, and STS-B), as well as pairwise inference tasks (MNLI, RTE, and QNLI).
We use the default train/dev/test split. The hyper-parameters are chosen based on the validation set (refer to the appendix for details).
After the model is trained, we make predictions on the test data and send the results to GLUE online evaluation service\footnote{https://gluebenchmark.com} to obtain final evaluation scores.

The evaluation scores on all datasets in GLUE benchmark are illustrated in Table \ref{tab:glue}. The performances of BERT-Base, BERT-Large, RoBERTa-Base, RoBERTa-Large, and T5-Large are reproduced using the official checkpoint provided by respective authors.
We only use self-contained constituency trees for the SST-2 dataset while other datasets are processed by Stanford parser\footnote{https://nlp.stanford.edu/software/lex-parser.shtml} to extract both dependency trees and constituency trees. For a fair comparison, all results of baseline models are reproduced by our own, which are close to the reported results.

As shown in the table, syntax-enhanced models always outperform corresponding baseline models. Most notably, Syntax-RoBERTa-Base achieves an average GLUE score of 82.1, lifting 1.3 scores from a standard RoBERTa-Base with the same setting. This is impressive as only a few extra parameters are introduced to the baseline model. Particularly, the improvements on CoLA and SST-2 datasets are fairly large, showing the generalization capability of Syntax-BERT and Syntax-RoBERTa on smaller datasets. Even on T5-Large, which is trained on more data and holds more advanced performances, our approach still outperforms the base model marginally (statistically significant under 4.3 p-value using paired t-test). We can see that more training data will improve the generalization capability of the model and compensate for the lack of syntax priors. On the other hand, syntactic information is useful in most cases, especially when training data or computation power is limited. 

\subsection{Ablation Study}
\label{sec:Ablation}
For a comprehensive understanding of the model design, we conduct ablation study with the following settings. (1) \textit{without topical attention}: the topical attention layer is removed, and a simple summation layer is replaced instead; (2) \textit{without syntax tree:} all the syntactic masks generated by the syntax trees are replaced by randomly generated masks, while the parameter size of the model remains unchanged; (3) \textit{without constituency/dependency tree}: only one kind of syntax tree is used in the model; (4) \textit{without parent / child / sibling / pairwise masks}: the corresponding masks are removed in the implementation.

\begin{table}[t]
 \centering
  {
  \small
  \begin{tabular}{l|c|c|c}
  \toprule
    \textbf{Model} & \textbf{SST-2} & \textbf{CoLA} & \textbf{STS-B}\\ 
    \midrule
    BERT-Large & 94.9 & 60.5 & 87.6/86.5\\
    \textbf{Syntax-BERT-Large} & \textbf{96.1} & \textbf{61.9} & \textbf{89.6/88.5}\\
    $~~$  w/o topical attention  & 95.1 & 61.6 & 88.4/87.3\\
    $~~$  w/o syntax trees & 95.0 & 60.5 & 88.0/87.1 \\
    $~~$  w/o dependency trees & 95.6 & 61.4 & 88.7/88.1\\
    $~~$  w/o constituency trees & 95.9 & 61.4 & 87.6/86.8\\
    $~~$  w/o parent masks & 95.5 & 60.9 & 88.7/87.2\\
    $~~$  w/o child masks & 95.3 & 61.2 & 88.3/86.8\\
    $~~$  w/o sibling masks & 95.8 & 61.5 & 89.0/88.1\\
    $~~$  w/o pairwise masks & - & - & 88.8/87.9\\
    \bottomrule
  \end{tabular}
  }
\caption{\small{Ablation study}}
\label{tab:Ablation}
\end{table}

As shown in Table \ref{tab:Ablation}, all datasets benefit from the usage of syntactic information. Generally, parent/child masks are of more importance than the sibling masks. Moreover, the topical attention layer is crucial to the performance of Syntax-BERT model, indicating the advantage of decomposing self-attention into different sub-networks and performing fine-grained aggregation. In addition, the pairwise mask is important on STS-B dataset and shows the benefit of cross-sentence attention.

\subsection{Structural Probe}
\label{sec:prob}
Our method ingests syntax trees into the model architecture directly. To examine if the representation learned by the model also captures syntactic knowledge effectively, we follow \citet{hewitt2019structural} to reconstruct a syntax tree of the entire sentence with linear transformation learned for the embedding space. If the syntax tree can be better reconstructed, the model is viewed to learn more syntactic information.
We evaluate the tree on \textit{undirected attachment score} -- the percent of undirected edges placed correctly, and \textit{Spearman correlation} between predicted and the actual distance between each word pair in a sentence. 
We probe models for their ability to capture the Stanford Dependencies formalism \cite{de2006generating}.
As shown in Table \ref{tab:probe}, for both metrics, the syntax-aware models get better scores than corresponding baseline models, indicating that Syntax-BERT is able to incorporate more syntax information than its vanilla counterparts.

\begin{table}[t]
 \centering
  {\small
  \begin{tabular}{l|c|c}
  \toprule
    \textbf{Model} & \textbf{UUAS} & \textbf{Spr.}  \\ 
    \midrule
    BERT-Base~\cite{devlin2018bert} & 79.8 & 0.85\\
    Syntax-BERT-Base & \textbf{81.1} & \textbf{0.88} \\
    \midrule
    BERT-Large~\cite{devlin2018bert} & 82.5 & 0.86\\
    Syntax-BERT-Large & \textbf{83.4} & \textbf{0.90}\\
    \midrule
    RoBERTa-Large~\cite{liu2020roberta} & 83.2 & 0.88\\
    Syntax-RoBERTa-Large & \textbf{84.6} & \textbf{0.93} \\
    \bottomrule
  \end{tabular}
  }
\caption{\small{The results of using Structural Probe to test whether different models contain syntactic information or not. UUAS denotes undirected attachment score, and Spr. denotes Spearman correlation.}}
\label{tab:probe}
\end{table}

\begin{figure}[t]
    \centering
    \includegraphics[width=\linewidth]{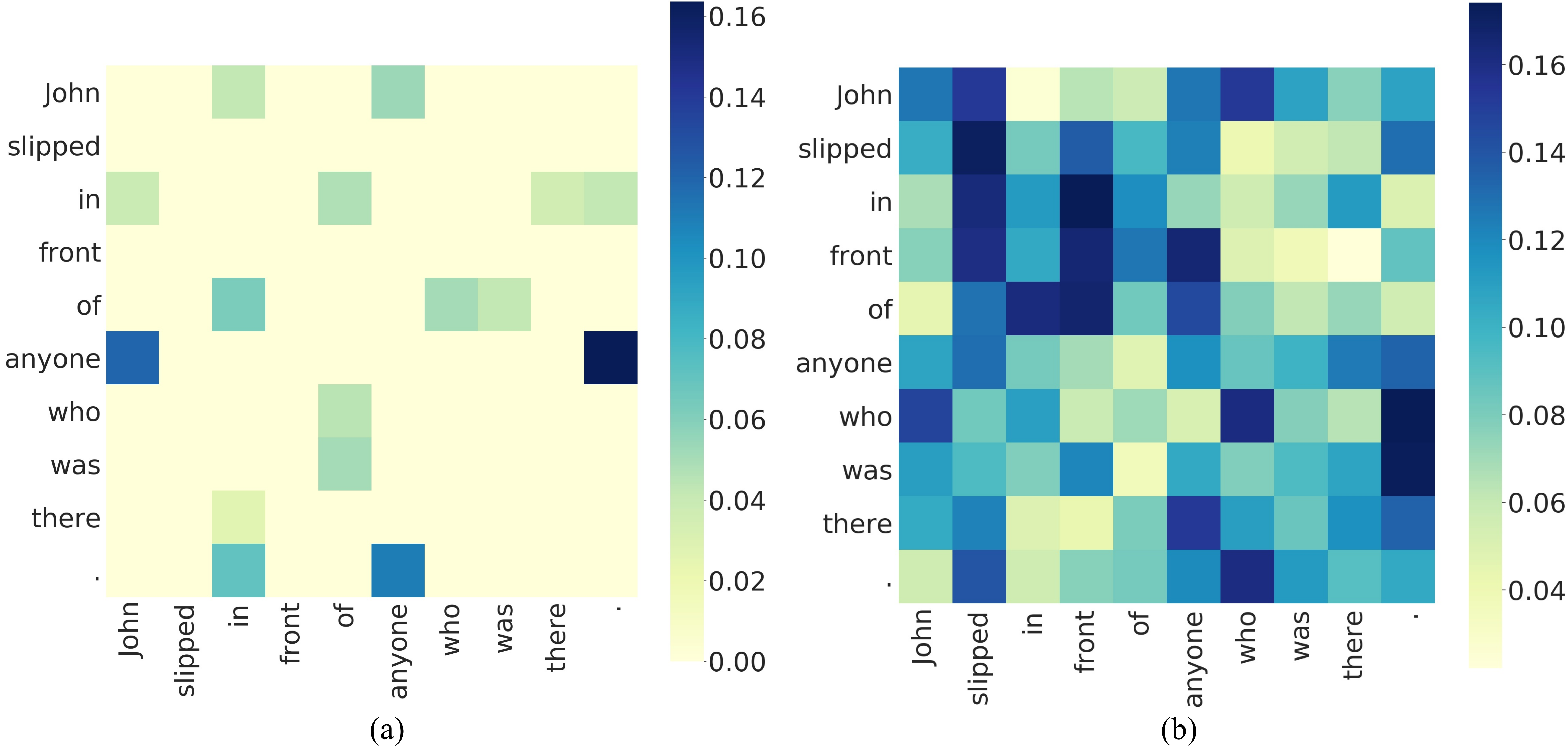}
    \caption{For an example sentence input, (a) The self-Attention scores of Syntax-Transformer corresponding to the sibling mask with $dist=3$. (b) The self-attention scores of a vanilla Transformer.}
    \label{fig:casestudy}
\end{figure}

\section{Discussion}

\subsection{Complexity analysis}
First, we choose BERT-Base as the base model to analyze the space complexity. As reported in \cite{devlin2018bert}, the number of trainable parameters in BERT-Base is about 110 million. Following \cite{strubell2018linguistically}, LISA-BERT-Base replaces one attention head in BERT-Base with a bi-affine attention head. Such an operation only adds a trainable matrix --- the bi-affine transformation matrix ---
in each layer, which brings about 0.6 million extra parameters.
Syntax-BERT-Base introduces a topical attention layer, which contains 1.0 million parameters in total for the BERT-Base version, while other parameters are inherited from vanilla BERT. Therefore, both LISA and Syntax-BERT add few parameters to the model and do not affect its original space complexity. 

We now analyze the time complexity of Syntax-BERT.
Assume the number of tokens in each sentence is $N$. 
First, constructing syntactic trees for each sentence and extract masking matrices can be prepossessed in the training phase or finish in $O(N^2)$ in the online inference phase. The time complexity of the embedding lookup layer is $O(N)$. Then, the attention score is calculated by $QK^\top\odot M$ with complexity $O(D_{Q}N^2)$, where $D_{Q}$ is the dimension of $Q$. Assume we have $M$ sub-networks. The complexity of masked self-attention is $O(MD_{Q}N^2)$. In the topical attention, the calculation process is very similar to traditional self-attention, only replacing $Q$ with a task-related vector. So it does not change the time complexity of BERT.
Finally, to get output representation, subsequent softmax and scalar-vector multiplication hold $O(D_{V}N)$ complexity, where $D_{V}$ is the dimension of $V$ for the topical attention. As such, the overall time complexity of Syntax-BERT is $O(N) + O(MD_{Q}N^2) + O(D_{V}N)= O(MD_{Q}N^2)$. When $M$ is small, the model has the same time complexity as vanilla BERT. Moreover, as the sub-networks are usually very sparse, the time complexity can be further improved to $O(MD_{Q}E)$ by a sparse implementation. Here $E \ll N^2$ denotes the average number of edges in a sub-network.

\subsection{Case Study}
We select the sentence ``John slipped in front of anyone who was there" in the CoLA dataset for case study. The task is to examine if a sentence conforms to English grammar. This sentence should be classified as negative since we use \textit{everyone} instead of \textit{anyone}. Syntax-Transformer classifies it correctly, but the vanilla transformer gives the wrong answer.

As visualized in Figure \ref{fig:casestudy}(a), the relationship between word pair \textit{(``anyone", ``.")} has been highlighted in one of the sub-networks, and the corresponding topical attention score for this sub-network in Syntax-Transformer is also very high. This shows a good explainability of Syntax-Transformer by correctly identifying the error term ``anyone", following a rule that ``anyone" is seldom matched with the punctuation ``.". However, a vanilla Transformer shows less meaningful self-attention scores, as illustrated in Figure \ref{fig:casestudy}(b). We give a briefing here, and please refer to the appendix for a complete description.

\section{Conclusion}
In this paper, we present Syntax-BERT, one of the first attempts to incorporate inductive bias of syntax trees to pre-trained Transformer models like BERT. The proposed framework can be easily plugged into an arbitrary pre-trained checkpoint, which underlines the most relevant syntactic knowledge automatically for each downstream task. We evaluate Syntax-BERT on various model backbones, including BERT, RoBERTa, and T5. The empirical results verify the effectiveness of this framework and the usefulness of syntax trees. In the future, we would like to investigate the performance of Syntax-BERT by applying it directly to the large-scale pre-training phase. Moreover, we are aiming to exploit more syntactic and semantic knowledge, including relation types from a dependency parser and concepts from a knowledge graph. 


\appendix

\section{Detailed settings}
Here we provide detailed settings for reproduction. The open-source code will be released when this paper is officially published. 

\subsection{Stanford Sentiment Treebank}
For raw Transformers, the number of layers is set as 12 and hidden dimension for each intermediate layer is set as 512. The probability of dropout is 0.1, and the hidden dimension of the final fully-connected layer is 2000. The word embedding vectors are initialized by GloVe (glove.840B.300d\footnote{https://nlp.stanford.edu/projects/glove/})~\cite{pennington2014glove} and fine-tuned during training. We use Adam optimizer with an initial learning rate 1e-4.

\subsection{Natural Language Inference}
For raw Transformers, we set layer number as 12, the hidden dimension of intermediate layers as 512, dropout ratio as 0.15, and the dimension of fully connected layer before Softmax activation as 2000. 
Learning rate is initialized as 5e-4, and Adam optimizer is used along with exponential learning rate decay of 0.9.

\section{Connection to GNN}
\label{sec:GCN}

A Transformer layer can be viewed as a special kind of Graph Neural Network (GNN), where each node represents for a word and all nodes construct a complete graph. To improve training speed and generalization ability, there are some previous works that advocate sparse architectures. For instance, Sparse Transformer~\cite{child2019generating} separates the full self-attention operation across several steps of attention for image classification. Star-Transformer~\cite{guo2019star} sparsifies the architecture by shaping the fully-connected network into a star-shaped structure consisting of ring connections and radical connections. 
In the architecture of Syntax-BERT, we also introduce sparsity to the complete graph network by decomposing it into multiple sub-networks. 
The most salient part of our approach is that the inductive bias is designed by syntax tree, which is a crucial prior for NLP tasks. In addition, as shown previously in Table \ref{tab:Ablation}, a random decomposition of the network also result in moderate performance enhancement. Similar phenomena is also reported in the image classification scenario with Graph Convolutional Network (GCN)~\cite{gurel2019anatomy}.

\end{document}